\documentclass[11pt]{article}
\usepackage{coling2020}
\usepackage{times}
\usepackage{url}
\usepackage{latexsym}

\usepackage{graphicx}
\usepackage{url}
\usepackage{hyperref}
\usepackage{amsmath}
\usepackage{amsfonts}
\usepackage{enumitem}
\usepackage{multirow}
\usepackage{hhline}
\usepackage{tikz}
\usepackage{xcolor}

\usepackage{tablefootnote}
\usepackage{booktabs}

\usepackage{microtype} 

\title{Understanding Pure Character-Based Neural Machine Translation: \\[1mm]The Case of Translating Finnish into English}

\author{Gongbo Tang$^1$\quad Rico Sennrich$^{2,3}$\quad Joakim Nivre$^1$ \medskip\\
  $^1$Department of Linguistics and Philology, Uppsala University\\
  $^2$Institute of Computational Linguistics, University of Zurich\\
  $^3$School of Informatics, University of Edinburgh\\
  {\tt firstname.lastname@\{lingfil.uu.se, ed.ac.uk\}}}
\date{}

\begin{document}
\maketitle
\begin{abstract}
Recent work has shown that deeper character-based neural machine translation (NMT) models can outperform subword-based models. However, it is still unclear what makes deeper character-based models successful. In this paper, we conduct an investigation into pure character-based models in the case of translating Finnish into English, including exploring the ability to learn word senses and morphological inflections and the attention mechanism. We demonstrate that word-level information is distributed over the entire character sequence rather than over a single character, and characters at different positions play different roles in learning linguistic knowledge. In addition, character-based models need more layers to encode word senses which explains why only deeper models outperform subword-based models. The attention distribution pattern shows that separators attract a lot of attention and we explore a sparse word-level attention to enforce character hidden states to capture the full word-level information. Experimental results show that the word-level attention with a single head results in 1.2 BLEU points drop. 
\end{abstract}

\section{Introduction}

\blfootnote{This work is licensed under a Creative Commons Attribution 4.0 International License. License
details: \url{http://creativecommons.org/licenses/by/4.0/}.}

Neural machine translation (NMT) has boosted machine translation significantly in recent years \cite{kal2013recurrent,cho2014learning,sutskever2014sequence,bahdanau15joint,luong2015effective,gehring2017convolutional,vaswani2017Attention}. However, it is still unclear how NMT models work due to the black-box nature of neural networks. Better understandings of NMT models could guide us in improving NMT systems. Currently most of the studies towards understanding NMT models only take into account subword-based (e.g.\ BPE-based) models. 
Deeper character-based (CHAR) models have been shown to perform better than BPE-based models \cite{cherry2018revisiting}. In this paper, we try to investigate the working mechanism of CHAR models. We explore the ability of CHAR models to learn word senses and morphological inflections and the attention mechanism. 

Previous studies have tried to interpret and understand NMT models by interpreting attention weights \cite{ghader2017what,raganato2018analysis,tang2018an,tang2019encoders}, using gradients \cite{he2019towards}, applying layer-wise relevance propagation \cite{ding2017visualizing}, probing classification tasks \cite{belinkov2017evaluating,belinkov2017what,Belinkov2020linguistic,Poliak2018on,tang2019encoders}, and more intrinsic analysis \cite{ghader2019intrinsic,voita2019bottom}. 
However, only \newcite{belinkov2017what,Belinkov2020linguistic} have probed character-based representations. \newcite{belinkov2017what} have only explored character-aware word-level representations, while we investigate fully character-level representations, which are also studied in \newcite{Belinkov2020linguistic}. We apply more composition methods to explore how CHAR models learn linguistic knowledge and how attention extracts features directly from characters. 

Probing classification tasks \cite{belinkov2017what} have emerged as a popular method to interpret the internal representations from neural networks. Given a probing classifier, the input is usually the representation of a word and the output is the corresponding linguistic tag. 
CHAR models pose new challenges for interpretability, and we investigate whether we can probe CHAR models in a way similar to (sub)word-based models. In addition, can we extract word sense and morphological information about the full word from individual hidden states, or is this information distributed across multiple states? This has implications for interpreting neural CHAR models, but can also inform novel architectures, such as sparse attention mechanisms. 
Thus we first investigate the ability of CHAR models to learn word senses and morphology in Section \ref{sec:ablility_linguistics}. 
We apply different methods to compose information from characters and demonstrate that the word-level information is distributed over all the characters but characters at different positions play different roles in learning linguistic knowledge. 
We also explore the effect of encoder depth to answer why CHAR models outperform BPE-based models only when they have the settings with deeper encoder. The probing results show that CHAR models need more layers to learn word senses. 
Then in Section \ref{sec:attention_distribution}, we move on to explore the attention mechanism. The distribution pattern shows that separators attract much more attention compared to other characters. 
To study the effect of enforcing characters to capture the full word-level information, we investigate a sparse attention mechanism, i.e. a model that only attends to separators, which can be viewed as a word-level attention. The BLEU score drops 1.2 points when we apply the word-level sparse attention. This implies that only attending to separators by a single attention head is workable but not enough to extract all the necessary information. 

The main findings are summarized as follows: 
\begin{itemize}[noitemsep,topsep=0pt]
  \item Word sense and morphological information is distributed over all the characters, but characters at different positions play different roles in learning linguistic knowledge. 
  \item CHAR models need more layers to encode word senses, which explains why only deeper models outperform BPE-based models. 
  \item Separators attract much more attention compared to other characters; we find that only attending to separators with a single head attention is workable but not sufficient for translation. 
\end{itemize}

\section{Experiments}

As RNN-based/CHAR models can in principle achieve state-of-the-art performance in NMT \cite{chen2018both,cherry2018revisiting} and most of analysis of NMT models are based on BPE-based models, we are interested in analyzing the working mechanisms of pure RNN-based CHAR models.\footnote{Also, training Transformer models at the character level with a large batch size requires enormous amounts of memory.} We follow \newcite{cherry2018revisiting} in using RNN-based models, and we focus on Finnish$\rightarrow$English (FI$\rightarrow$EN), because training CHAR models requires huge computational resources. 

We first train CHAR models with different encoder depths and a BPE-based model for comparison. Then we explore how CHAR models learn word senses and morphological inflections via probing classification tasks, using representations generated by the trained models. 
For the morphological probing tasks, the classifiers predict the morphological tag given the representation of a token. 
For the word sense disambiguation (WSD) probing task where learning word senses is needed, the input to the classifier and the output are different from the classifiers in the morphological probing tasks. Instead, the representations of an ambiguous word and its candidate translation are both fed into the classifier and then the classifier predicts whether the candidate translation is correct or not. 

\subsection{Data}
We train NMT models on the WMT15 shared task data \cite{wmt15} for FI$\rightarrow$EN to be able to compare with \newcite{cherry2018revisiting}. There are about 2.1M sentence pairs in the training set after preprocessing with Moses scripts. 

For the WSD probing task, we use the FI--EN part of the \textit{MuCoW} \cite{raganato2019mucow} test set, which is a multilingual test suite for WSD in the WMT19 shared task. It has 2,117 annotated sentences. Each annotation provides the ambiguous Finnish word, the domain of the sentence, and a set of translation candidates of the ambiguous word including both correct and incorrect translations. For each ambiguous word from an annotation, we generate multiple instances that are labeled with one translation candidate and a binary value indicating whether it corresponds to the correct sense. 1,000/1,000 instances are randomly selected as the development/test sets, and the remaining 6,325 instances are used for training the probing classifiers.\footnote{Note that \textit{MuCoW} has both in-domain (news and books) and out-of-domain data (subtitles), and the results are only based on the in-domain data unless otherwise specified. The ratio of in-domain data and out-of-domain data is around 1:2.} 

\begin{table}[tbp]
\centering
\scalebox{0.9}{
\begin{tabular}{lccccccc}
\toprule
Feature&\ &POS& Grammatical case& Locative case& Number& Infinitive& Voice \\
\midrule
\# Tag & \ &16& 4& 8& 8& 3& 2 \\
Training& \  &11,678& 5,147& 1,495& 7,603& 301& 1,723 \\
Dev/Test& \  &\phantom{0}1,500& \phantom{0,}650& \phantom{0,}200& 1,000& \phantom{0}40& \phantom{0,}200 \\
\bottomrule
\end{tabular}}
\caption{\label{table-data-morph-stat} Statistics of in-domain data from \textit{MuCoW} for morphological probings. }
\end{table}

To extend \textit{MuCoW} for the morphological probing tasks, we use the RNNTagger,\footnote{\url{https://www.cis.uni-muenchen.de/~schmid/tools/RNNTagger/}} which is trained on FinnTreebank2\footnote{\url{http://www.ling.helsinki.fi/kieliteknologia/tutkimus/treebank/index.shtml}} to generate the morphological tags. 
Finnish is a morphologically rich language, and in addition to \textit{POS}, we generate data for 5 other morphological features: \textit{grammatical case, locative case, number, infinitive}, and \textit{voice}. 
These features vary in the types of tag. The data is roughly split into training/development/test sets at the ratio of 8:1:1. Each data entry for the probing tasks contains the representation of a token and the morphological tag. The detailed statistics are provided in Table~\ref{table-data-morph-stat}. 

\subsection{Experimental Settings}

\paragraph{NMT models} 
We use the \textit{Sockeye} \cite{Hieber2017sockeye} toolkit to train NMT models. The encoder is a stack of 1 bidirectional RNN and 6 unidirectional RNNs, and the decoder has 8 unidirectional RNNs. We choose long short-term memory (LSTM) RNN unit \cite{hochreiter1997long}. The size of embeddings and hidden units is 512. We tie the source, target, and output embeddings. 
The beam size is 8 during inference. We employ the models that have the best perplexity on the validation set for evaluation. BLEU scores \cite{papineni2002bleu} are computed by \textit{sacrebleu} \cite{post2018sacre}. 
For CHAR models, we add separators between any two tokens including punctuation marks, and input character sequences to the model directly. The character vocabulary size is 379. For the BPE-based model, we learn a joint BPE model with 32K subwords \cite{sennrich16sub}. 
As \newcite{cherry2018revisiting} have shown that the depth is crucial to the success of CHAR models, we train a 4-layer CHAR model to study the effect of depth.

\paragraph{Probing classifiers}

These probing classifiers are feed-forward neural networks with only one hidden layer, using ReLU non-linear activation. The size of the hidden layer is set to 512. We use the Adam learning algorithm \cite{Kingma2014AdamAM}. The classifiers are trained using a cross-entropy loss. Each classifier is trained for 180/100 epochs in the WSD/morphological probing tasks and the one that performs best on the development set is selected for evaluation. We train 5 times with different seeds for each classifier and report average accuracy. 

In contrast to word-level hidden states, a word consists of multiple character-level hidden states in CHAR models. We are interested in how the word-level information, including word senses and morphological inflections, is distributed over the character hidden states, in a single state or spread over all hidden states. Thus we explore the following methods for composition: 
\begin{itemize}[noitemsep,topsep=0pt]
  \item mean pooling: mean of hidden states 
  \item max pooling: max of hidden states in each dimension
  \item last pooling: last hidden state
  \item first pooling: first hidden state
  \item randLSTM: output of a randomly initialized LSTM, whose input are the hidden states of characters; we use a 1-layer bidirectional LSTM with parameters initialized uniformly at random from $[-\frac{1}{\sqrt{d}},\frac{1}{\sqrt{d}}]$, where $d$ is the hidden size of the LSTM \cite{Wieting2019NoTR}.\footnote{Note that we first use randLSTM to generate the composition, then we feed the output into the following classifiers. That is to say, the parameters of the randLSTM are fixed during the training of the following probing classifiers. This can reduce the effect of LSTMs on the probing classifiers. }   
\end{itemize}

\begin{table}[htbp]
\centering
\begin{tabular}{lrrrr}
\toprule
System & \newcite{cherry2018revisiting}& bpe-d7&char-d4  &char-d7 \\
\midrule
Encoder Depth  & 6& 7&4& 7\\
BLEU & 19.5 & 16.9 &16.3& 17.2\\
\bottomrule
\end{tabular}
\caption{\label{table-bleu} BLEU scores of the NMT models on FI$\rightarrow$EN. }
\end{table}

\begin{table}[ht!]
\centering
\scalebox{0.66}{
\begin{tabular}{cl|ccc|ccccc|ccc|ccc}
\toprule
\multirow{2}{*}{Feature} &model&\multicolumn{3}{c|}{char-d7}&\multicolumn{3}{c}{bpe-d7}& \multirow{2}{*}{Feature} &model&\multicolumn{3}{c|}{char-d7}&\multicolumn{3}{c}{bpe-d7}\\
\cmidrule(l){2-8} \cmidrule(l){10-16}
&layer&1 & 3 & 5 & 1 & 3 & 5&\ &layer&1 & 3 & 5 & 1 & 3 & 5\\
\midrule
\multirow{5}{*}{WSD}&mean&71.95 &79.94 &81.57 &83.02 &84.53 &83.77 &\multirow{5}{*}{POS}&mean &93.29 &93.91 & \textbf{94.80} &41.29 &42.07 &40.79 \\
&max&71.76 &74.91 &75.60 &84.65 &85.28 &83.52 & &max &92.23 &93.56 &94.20 &43.35 & \textbf{44.35} &42.31 \\
&last&74.78 &80.25 &80.88 &82.58 &84.59 &84.40 & &last &73.81 &81.37 &87.43 &40.08 &39.83 &39.43 \\
&first&74.28 &81.89 &82.96 &84.40 &\textbf{85.66}&83.65 & &first &92.08 &91.92 &91.43 &39.28 &40.91 &39.19 \\
&randLSTM&74.65 &\textbf{83.90}&82.89 &82.26 &80.44 &81.07 & &randLSTM &92.57 &92.83 &92.56 &35.91 &35.52 &35.92 \\
\addlinespace
\addlinespace
\multirow{5}{*}{Grammatical}&mean &60.40 &58.37 &58.15 &66.03 &64.12 &61.75 &\multirow{5}{*}{Locative}&mean &26.00 &25.60 &27.30 &47.90 &45.40 &40.30 \\
&max &59.97 &57.94 &58.28 & \textbf{67.14} &63.23 &60.03 & &max &26.20 &26.80 &24.60 &49.30 &46.60 &41.20 \\
&last &57.94 &54.68 &56.55 &66.62 &64.09 &61.35 & &last &24.40 &21.30 &24.50 & \textbf{50.70} &47.10 &40.10 \\
&first &55.78 &54.06 &55.54 &65.66 &63.63 &59.29 & &first &24.00 &23.40 &20.80 &48.40 &42.00 &34.40 \\
&randLSTM & \textbf{92.25} &90.62 &87.51 &55.88 &56.34 &53.66 & &randLSTM & \textbf{81.60} &75.10 &78.30 &30.70 &32.00 &27.80 \\
\addlinespace
\addlinespace
\multirow{5}{*}{Number}&mean &75.10 &75.26 &75.34 &72.38 & \textbf{73.54} &70.98&\multirow{5}{*}{Infinitive}&mean &53.50 &57.00 &58.50 &55.50 &53.50 &51.50 \\
&max &73.92 &73.90 &73.50 &72.82 &73.02 &71.10 &&max &58.50 &52.50 &54.00 &59.00 &57.50 & \textbf{61.50} \\
&last &71.90 &72.04 &72.10 &72.54 &72.40 &70.56 &&last &56.50 &57.00 &53.00 &58.00 &55.50 &51.50 \\
&first &72.94 &73.06 &72.90 &72.60 &72.38 &69.88 &&first &50.00 &47.50 &49.00 &57.50 &52.50 &54.00 \\
&randLSTM &95.32 & \textbf{95.52} &94.12 &70.10 &69.68 &70.00 &&randLSTM & \textbf{80.00} &75.00 &75.00 &55.00 &55.00 &55.00 \\
\addlinespace
\addlinespace
\multirow{5}{*}{Voice}&mean &84.00 &83.70 &84.30 &85.60 &85.70 &84.20 \\
&max &84.00 &84.00 &84.00 &85.80 & \textbf{86.00} &85.20 \\
&last &82.40 &84.10 &83.10 &83.80 &83.60 &83.00\\
&first &81.10 &83.60 &83.10 &84.60 &84.80 &83.40\\
&randLSTM &94.50 &95.00 & \textbf{95.20} &84.00 &82.90 &83.00\\
\bottomrule
\end{tabular}
}
\caption{\label{table-result} Accuracy (\%) on WSD and morphological probing tasks using hidden states from \textit{char-d7} and \textit{bpe-d7}, with different composition methods. The numbers in bold are the best accuracy of each model in each probing task. The results of \textit{char-d7} on morphological probing tasks consider separators as the last character of a word while those on the WSD probing tasks do not take separators into account.}
\end{table}

\noindent
The \textit{mean} pooling method, which simply averages all the hidden states of a word, can tell us how much word sense or morphological information has been encoded into the word and serves as the baseline. The \textit{first/last} pooling method detects how much word sense or morphological information can be captured by a single character. 
\textit{randLSTM} can test whether word sense or morphological information need to be modeled by a more complicated composition method or has been encoded into each hidden state and only need a simple \textit{mean} pooling composition. 
Note that we are not pursuing better composition methods for probing tasks but investigating how CHAR models encode the word sense and morphological information into hidden states of characters. 
We also apply these composition methods to subwords.

\subsection{Results}

\subsubsection{BLEU Scores of NMT Models}
Table~\ref{table-bleu} gives the BLEU scores of the three NMT models, which are used for the following investigation. In accordance with \newcite{cherry2018revisiting}, our deeper CHAR model (\textit{char-d7}) indeed outperforms the BPE-based model with the same number of layers (\textit{bpe-d7}). The CHAR model with 4 layers (\textit{char-d4}) is inferior to the other two models as expected. 
The result of \newcite{cherry2018revisiting} in Table \ref{table-bleu} is obtained using 6 bidirectional gated recurrent units (GRUs) \cite{cho2014learning} in the encoder, which are 12 unidirectional LSTMs. Our encoder only has 1 bidirectional LSTM and 6 unidirectional LSTMs. In addition, we do not apply label smoothing technique in our models. We assume that these differences in settings cause the performance gap. 
Nevertheless, we focus on exploring how CHAR models work rather than pursuing better performance.

\subsubsection{Accuracy in Probing Tasks}
\label{ssub:results_prob}

Table~\ref{table-result} gives the accuracy in the WSD probing task and different morphological probing tasks, using hidden states from \textit{char-d7} and \textit{bpe-d7} with different composition methods. Results of \textit{bpe-d7} are given for comparison. The bold numbers are the best accuracy of each model in each probing task. 

\begin{figure}[htbp]
\begin{minipage}[t]{0.48\linewidth}
    \begin{center}
    \includegraphics[totalheight=5cm]{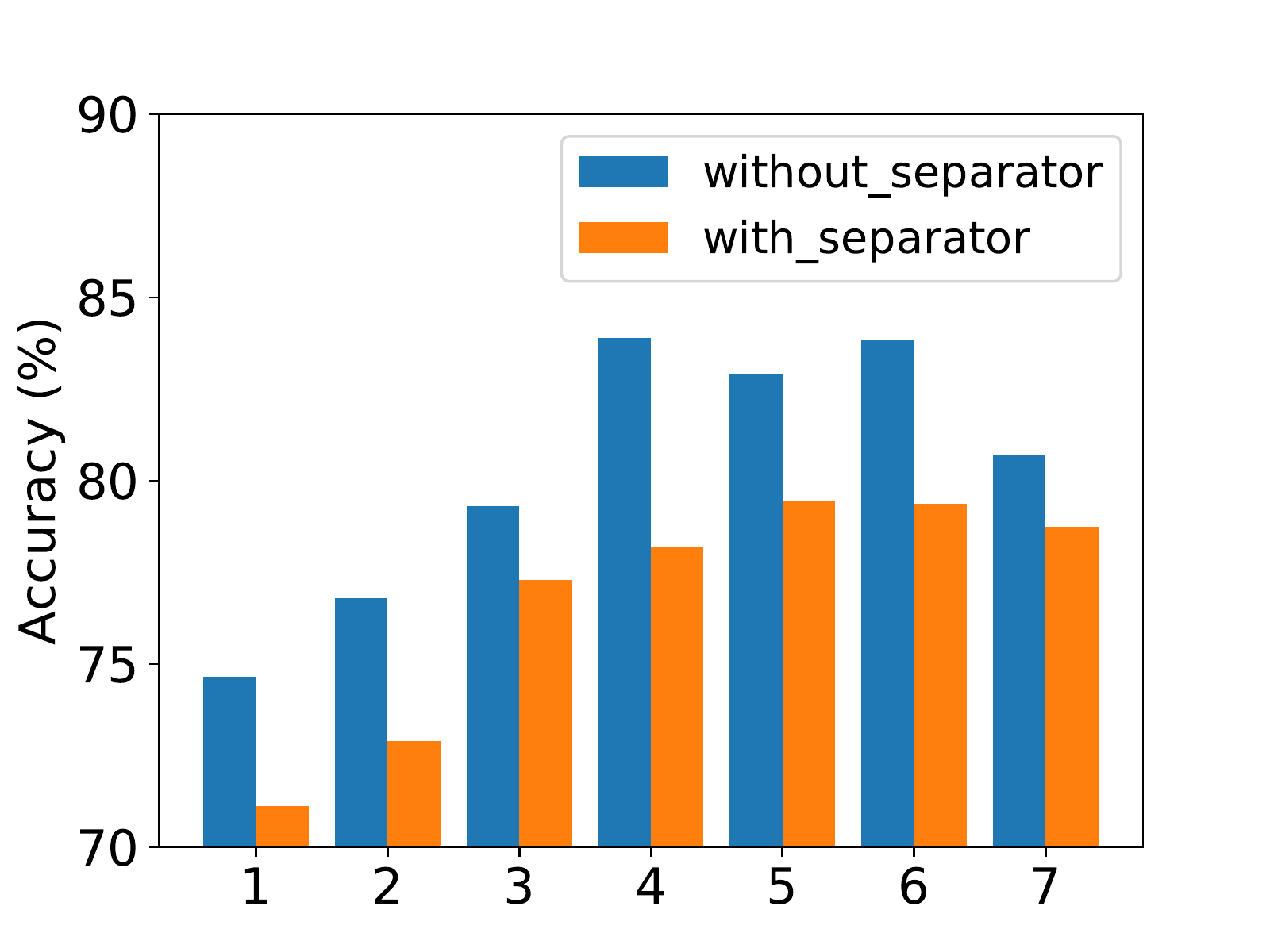}
    \caption{Accuracy of each layer from \textit{char-d7} on the WSD probing task, considering separators or not, using \textit{randLSTM} for composition.}
    \label{fig:wsd_separator}
    \end{center}
\end{minipage}%
    \hfill%
\begin{minipage}[t]{0.48\linewidth}
    \begin{center}
    \includegraphics[totalheight=4.5cm]{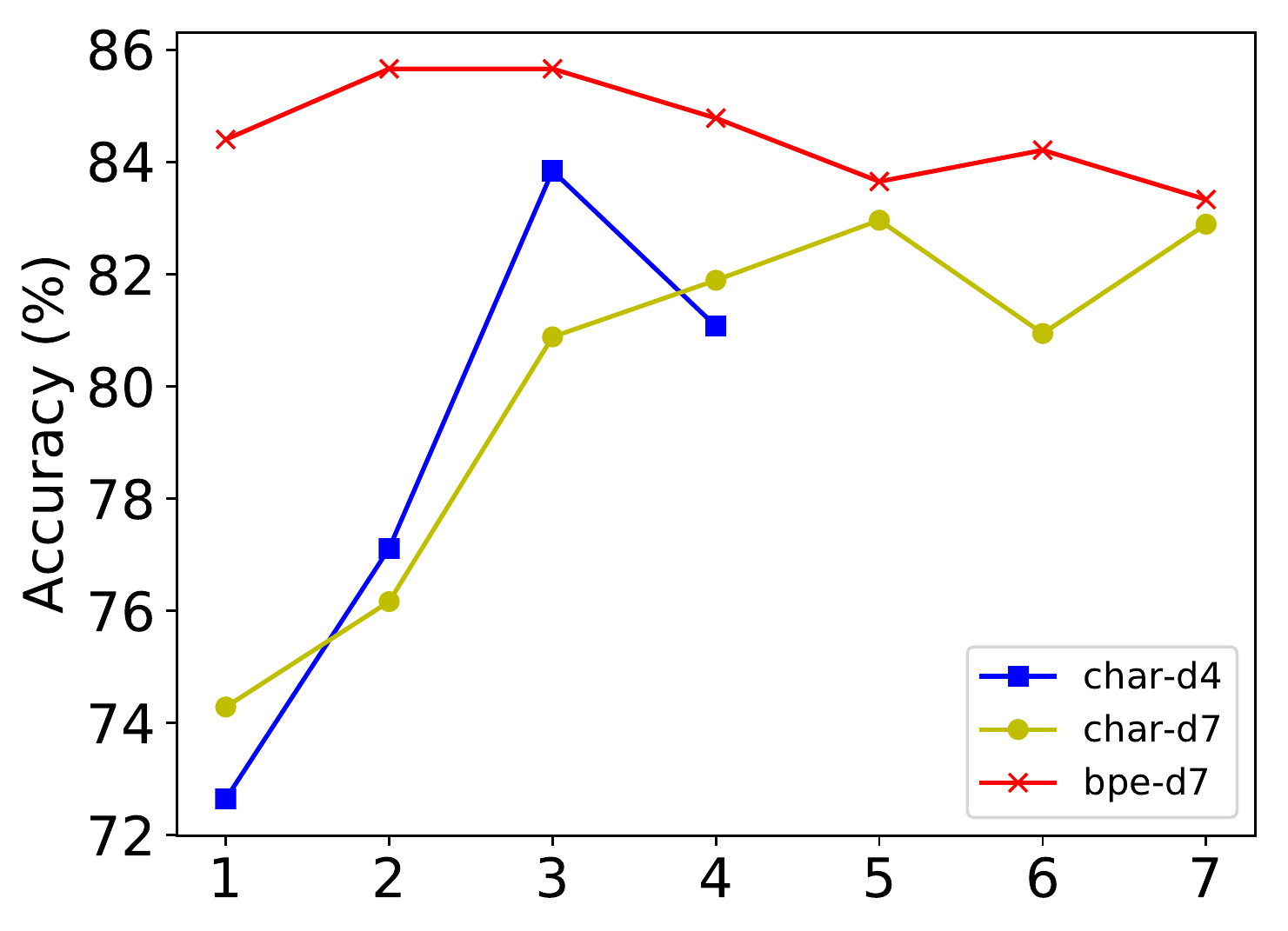}
    \caption{Accuracy evolution over layers on the WSD probing task, using \textit{first} pooling for composition. }
    \label{fig:wsd_evolution}
    \end{center}
\end{minipage} 
\end{figure}

For the WSD probing task, we can see that using the first hidden state of the ambiguous words from the 3rd layer of \textit{bpe-d7} achieves the highest accuracy (85.66\%). In \textit{char-d7}, hidden states from higher layers tends to perform better than those from lower layers when using pooling methods for composition but not when using \textit{randLSTM} for composition.

We can tell that \textit{char-d7} achieves significantly better performance than \textit{bpe-d7} on all the morphological probing tasks, which is consistent with previous finding that character-level hidden states are better for learning morphology \cite{Lee2017fully,belinkov2017what,durrani2019one,Belinkov2020linguistic}. 
In the \textit{POS} probing task, the hidden states of characters are much superior to those of subwords.\footnote{We have noted the exceptionally low accuracy of \textit{bpe-d7}, but we have not found any errors from the perspective of experimental settings.} This indicates that the \textit{POS} information is better encoded into hidden states than the other morphological features. 
In the \textit{locative} probing task, although \textit{char-d7} using \textit{randLSTM} performs much better than \textit{bpe-d7}, the performance of pooling methods is far behind that of \textit{bpe-d7}. We interpret this as showing that \textit{locative} case information is not simply distributed over character hidden states and we need a more complicated composition method to extract the information. 
Both \textit{char-d7} and \textit{bpe-d7} perform well on the \textit{voice} probing task which is a relatively easy binary classification task.

\section{Learning Linguistics}
\label{sec:ablility_linguistics}

In this section, we interpret the ability of CHAR models to encode word senses and morphological inflections, by exploring the effect of separators, composition methods, the evolution over layers, and the robustness to domain-mismatch. We only analyze the encoder hidden states in the context of probing tasks. As the attention extracts features from encoder hidden states in a different way from our composition methods -- pooling and \textit{randLSTM}, the findings may not apply to the entire CHAR model.

\subsection{Learning Word Senses}
\label{sub:learn_wsd}

\subsubsection{The Effect of Separators}
Separators indicate word boundaries in the input sequences, which potentially are viewed as the end of a word by the CHAR model. We test the role of separators in learning word senses by testing the effect on the WSD probing task. 
Figure~\ref{fig:wsd_separator} displays the WSD accuracy of representations from different encoder layers with and without considering separators as the last character of a word, using \textit{randLSTM} for composition. We can see that considering separators results in a lower accuracy which means that separators have a negative effect on learning word senses when we compose all the characters. 
However, the information of characters from a word passes to separators in the encoder. Separators do carry some of the word sense information and they can achieve up to 79.8\% in accuracy. 
We speculate that separators also capture some morphological information which confuses the classifier in identifying word senses, and we will explore it in Section \ref{sub:learn_morph}. Note that the following results in this section do not consider separators. 

\begin{table}[t!]
\begin{center}
\begin{tabular}{lcc}
\toprule
Model & \textit{char-d7} & \textit{bpe-d7}\\ 
\midrule
in-domain&83.9\% &85.7\% \\
out-of-domain &45.7\% & 49.1\% \\
\bottomrule
\end{tabular}
\caption{\label{table-wsd-domain} Best accuracy of \textit{char-d7} and \textit{bpe-d7} on the WSD probing task, on in-domain and out-of-domain test sets. }
\end{center}
\end{table}

\subsubsection{Composition Methods}

Even though \textit{first} pooling only utilizes the first hidden state of a word, it performs better than \textit{mean} and \textit{max} which use all the hidden states. We can infer that this CHAR model encodes more word sense information into the first characters of words. 
However, \textit{randLSTM} can achieve higher accuracy than \textit{first}. We conclude that this CHAR model also distributes the word sense information to other characters but we need a more complicated composition method to extract more word sense information. 

For \textit{bpe-d7}, \textit{first} achieves the best accuracy among all the composition methods. We can tell that both the CHAR and BPE-based models encode much sense information into the first character/subword but the first subword is enough to represent the word sense. Moreover, \textit{randLSTM} performs worse than the simple pooling methods in \textit{bpe-d7}, which indicates that the information about word senses has been well represented by hidden states and we do not need a more complicated method to further extract it.

\subsubsection{Evolution over Layers}

Figure~\ref{fig:wsd_evolution} shows the accuracy evolution over layers of both CHAR and BPE-based models, using \textit{first} for composition. 
In the first layer, \textit{char-d4/7} performs much worse than \textit{bpe-d7}. However, the learning curve of CHAR models is much steeper than \textit{bpe-d7}, especially in the first three layers. 
In the 7th layer, \textit{char-d7} performs almost as well as \textit{bpe-d7}. 
We speculate that it takes several layers for the first character to learn the basic sense of a word and it need more layers to learn the contextualized/disambiguated word sense. 
This can explain why previous shallow CHAR models do not perform well, as word senses are learned layer by layer. 

\subsubsection{Robustness to Domain-Mismatch}

CHAR models have been shown robust to spelling mistakes, rare words, morphology, and compounds \cite{Lee2017fully,cherry2018revisiting}. The meaning of an ambiguous word is likely to vary with domains. Thus, here we investigate the robustness to domain-mismatch of CHAR models when learning word senses. We directly test the models trained on in-domain data using the out-of-domain test set. 

Table~\ref{table-wsd-domain} gives the best accuracy on in-domain and out-of-domain test sets. The accuracy of both models has a substantial drop on the out-of-domain test set which is consistent with the finding from \newcite{raganato2019mucow}. The drop of \textit{char-d7} is even bigger than that of \textit{bpe-d7} which indicates that CHAR models are not more robust to domain-mismatch when learning word senses compared to BPE-based models.

\subsection{Learning Morphology}
\label{sub:learn_morph}

\subsubsection{The Effect of Separators}

We have demonstrated that separators have a negative effect on the WSD probing task in Section \ref{sub:learn_wsd}. 
As separators indicate the word boundary information and morphological features are reflected over the middle parts or the last parts of a word in Finnish, here we hypothesize that these separators are important to the morphological probing tasks. 
We measure the effect of separators by comparing the performance on the morphological probing tasks whether considering separators as the last character of a word. 

\begin{figure}[htbp]
\begin{minipage}[t]{0.4\linewidth}
    \begin{center}
    \includegraphics[totalheight=4.2cm]{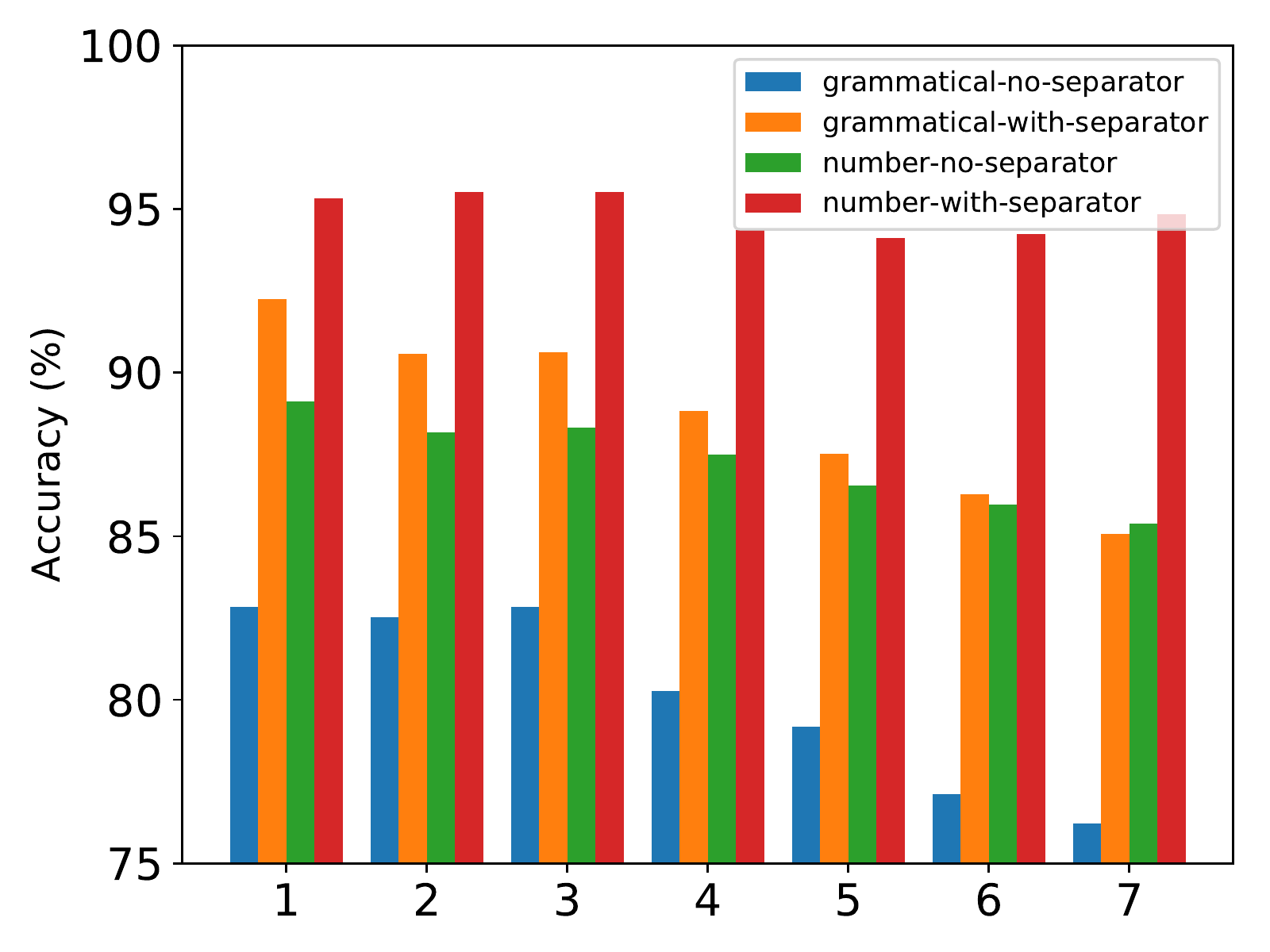}
    \caption{Accuracy on \textit{grammatical} and \textit{number} probing tasks, considering the separators or not, using \textit{randLSTM} for composition. }
    \label{fig:morp_separator}
    \end{center}
\end{minipage}%
    \hfill%
\begin{minipage}[t]{0.56\linewidth}
    \begin{center}
    \includegraphics[totalheight=3.9cm]{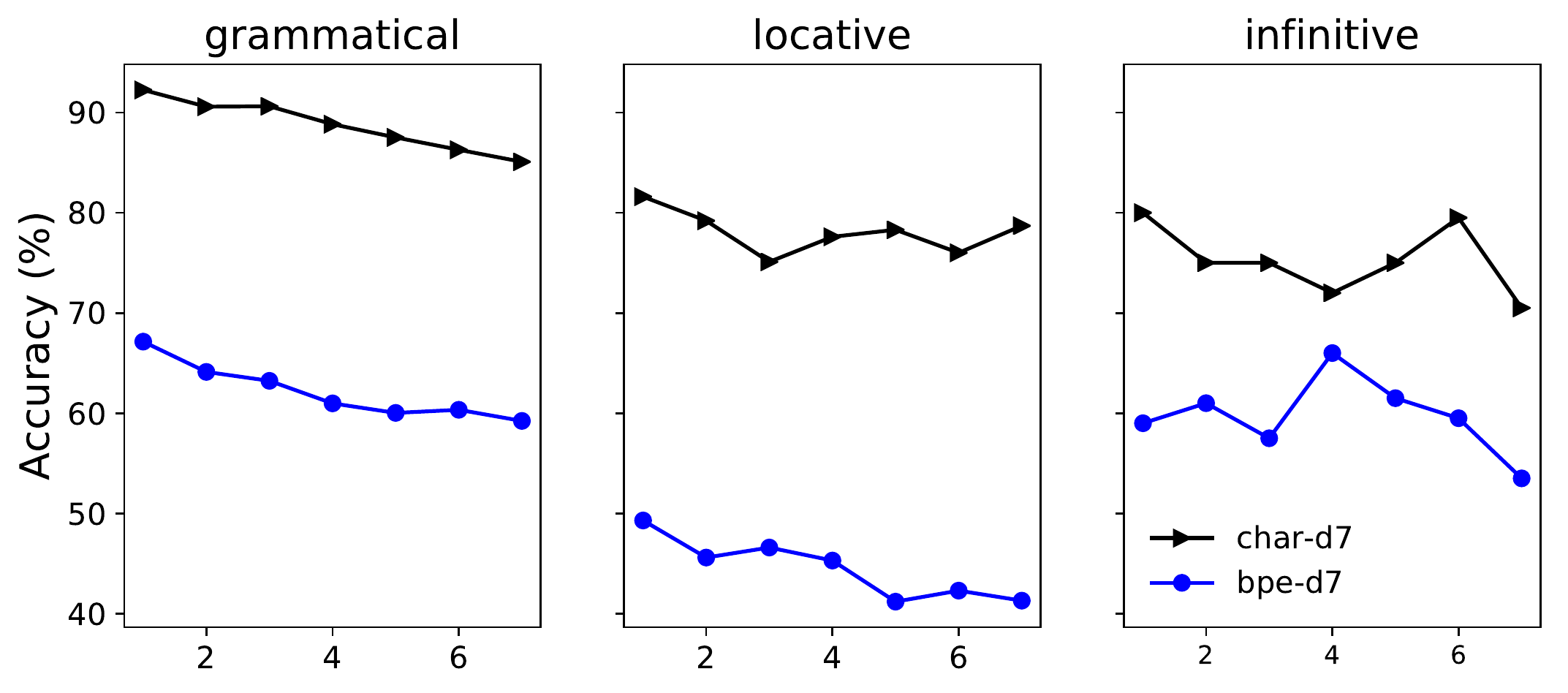}
    \caption{Accuracy evolution over layers on \textit{grammatical}, \textit{locative}, and \textit{infinitive} probing tasks, in \textit{char-d7} and \textit{bpe-d7}, using \textit{randLSTM} and \textit{max} for composition, respectively. }
    \label{fig:mor_trend}
    \end{center}
\end{minipage} 
\end{figure}

We find that the representations considering the separators are evidently superior. Figure~\ref{fig:morp_separator} displays the comparison on the \textit{grammatical} and \textit{number} probing tasks and we get the same pattern in the other morphological probing tasks. These results indicate that separators capture much of the word-level morphological information which is not encoded into other characters. The following results in this section take separators into account.

\subsubsection{Composition Methods}

In Table~\ref{table-result}, \textit{max} also considers all the hidden states but has different performance compared to \textit{mean}, especially on \textit{infinitive} and \textit{number}.  
\textit{first} and \textit{last} are inferior to \textit{mean} and \textit{max} which implies that the first and the last character of the word do not capture all the word-level morphological information, even though we have shown that the last character (separator) has some crucial information. 
In particular, \textit{last} only achieves 88.16\% on the POS probing task while \textit{mean} achieves 94.80\%. 
Thus, we conclude that the model has not learned to build a full morphological representation of words at the first or last position, but that the information remains distributed across positions. 
In \textit{bpe-d7}, \textit{max} pooling achieves the best results in 4 out of 6 probing tasks. \textit{first} and \textit{last} are usually inferior to \textit{mean/max}. 

\textit{randLSTM} performs significantly better than \textit{mean/max} except on the \textit{POS} probing task, even though the LSTM is randomly initialized without any training. The gaps vary from 10.9\% to 54.1\%, especially in the probing task on \textit{locative}. 
We can infer that the information of word structure is not well encoded into hidden states, thus we need a more complicated composition method to abstract the information. 

For \textit{POS}, \textit{randLSTM} performs worse than \textit{mean/max}. We attribute this to the fact that \textit{POS} is a global feature compared to other morphological features and has been well encoded into hidden states. Thus, it does not need further extraction. 
In contrast to the results for \textit{char-d7}, \textit{randLSTM} for \textit{bpe-d7} is not as good as the simple pooling methods in any of the probing tasks. We infer that the morphological information has been well encoded into subword hidden states and does not need further composition. 

\subsubsection{Evolution over Layers}

Figure~\ref{fig:mor_trend} exhibits the evolution of learning \textit{grammatical}, \textit{locative}, and \textit{infinitive} features over layers. We can see that the overall accuracy of \textit{char-d7} and \textit{bpe-d7} tends to go down over layers (also in other morphological probing tasks).  The results are consistent with the previous findings in \newcite{belinkov2017what,belinkov2017evaluating,Belinkov2020linguistic} that hidden states from the lower layers are better at learning morphology.

\begin{figure}[htbp]
\begin{minipage}[t]{0.48\linewidth}
    \begin{center}
    \includegraphics[totalheight=4.7cm]{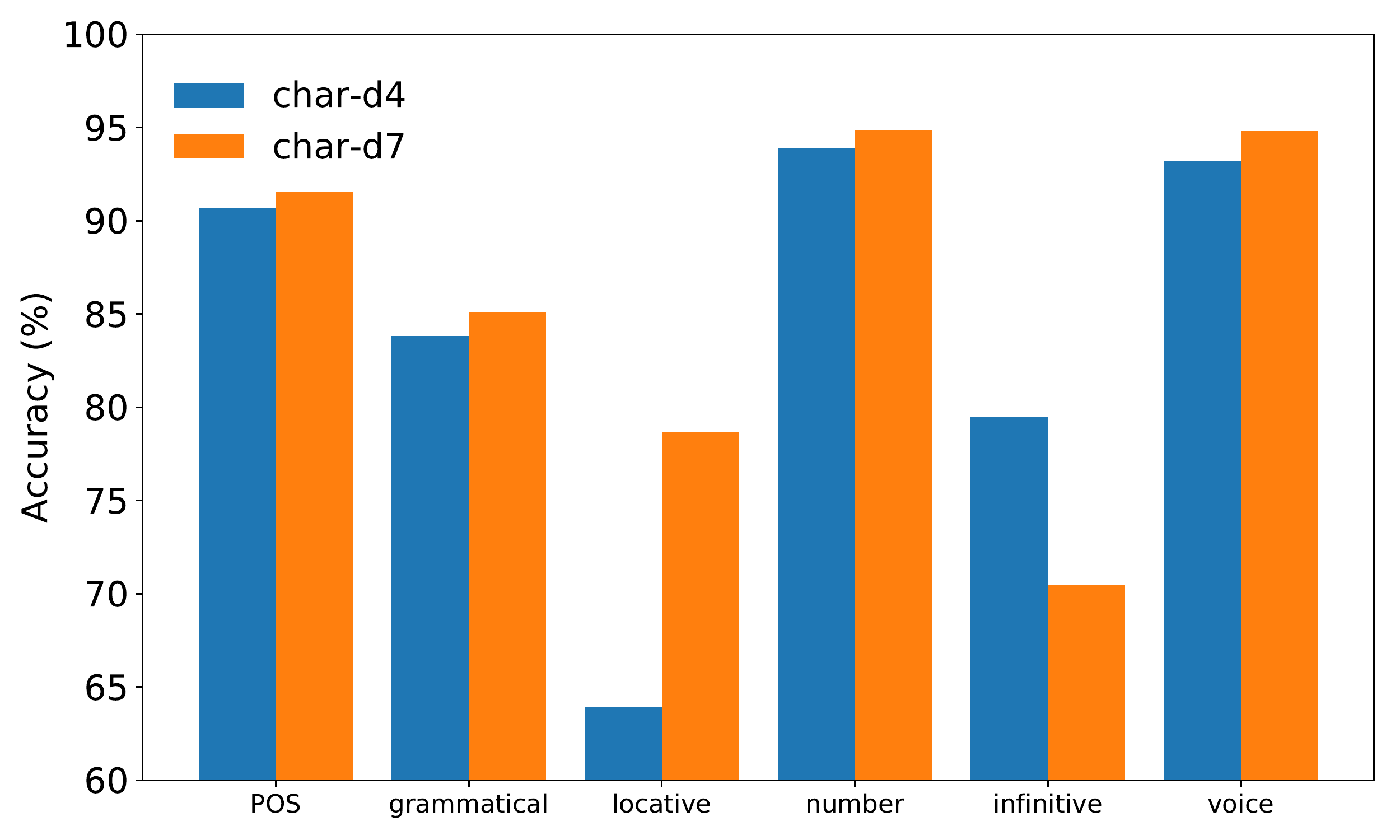}
    \caption{Accuracy on the six morphological probing tasks, using the last layer from \textit{char-d4} and \textit{char-d7}.}
    \label{fig:d4d7last}
    \end{center}
\end{minipage}%
    \hfill%
\begin{minipage}[t]{0.46\linewidth}
    \begin{center}
    \includegraphics[totalheight=4.52cm]{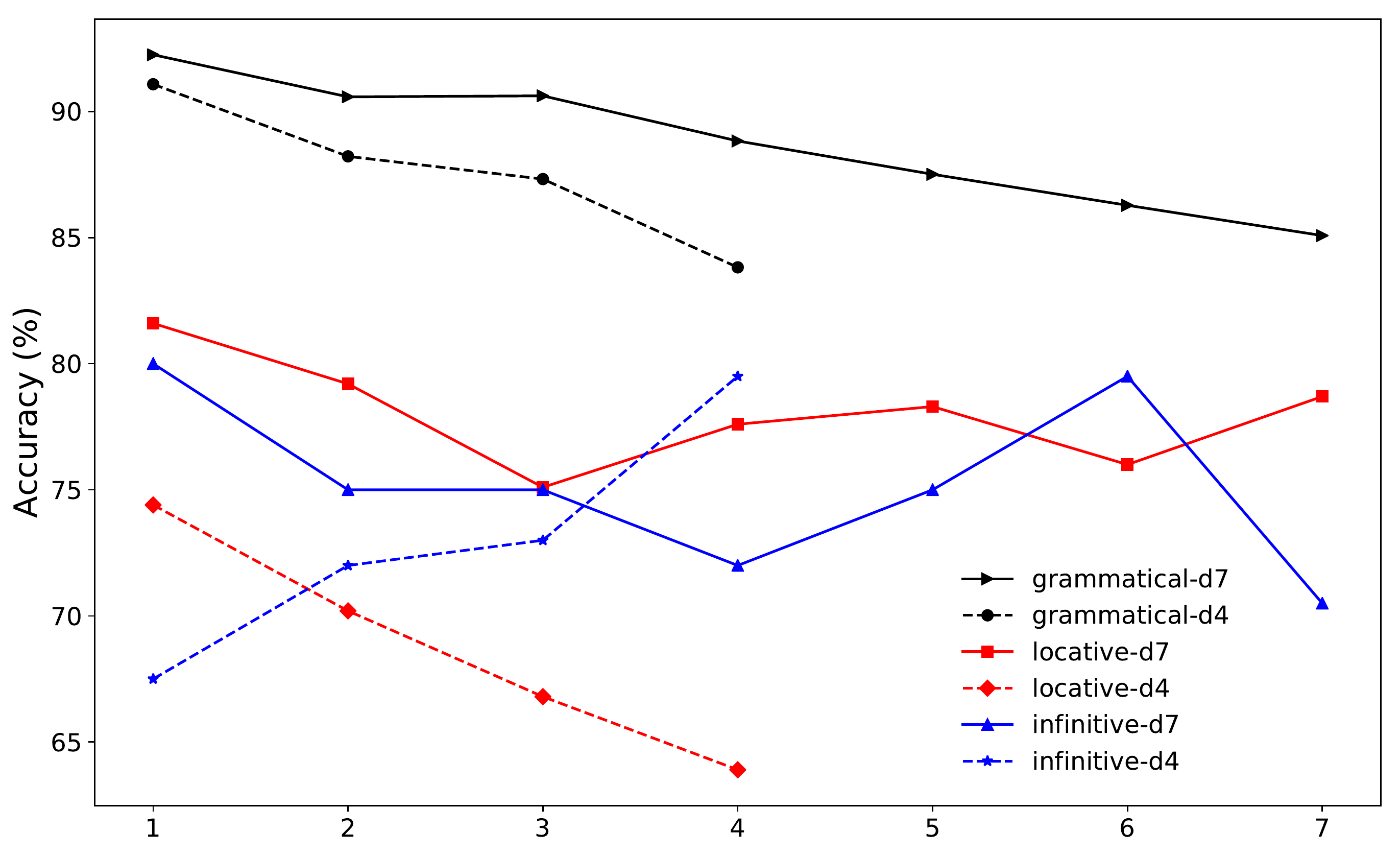}
    \caption{Accuracy evolution over layers of \textit{char-d4} and \textit{char-d7}, on \textit{grammatical}, \textit{locative}, and \textit{infinitive} probing tasks.} 
    \label{fig:d4d7evolution}
    \end{center}
\end{minipage} 
\end{figure}

\begin{figure}[h!]
\begin{minipage}[t]{0.51\linewidth}
    \begin{center}
    \includegraphics[totalheight=4.4cm]{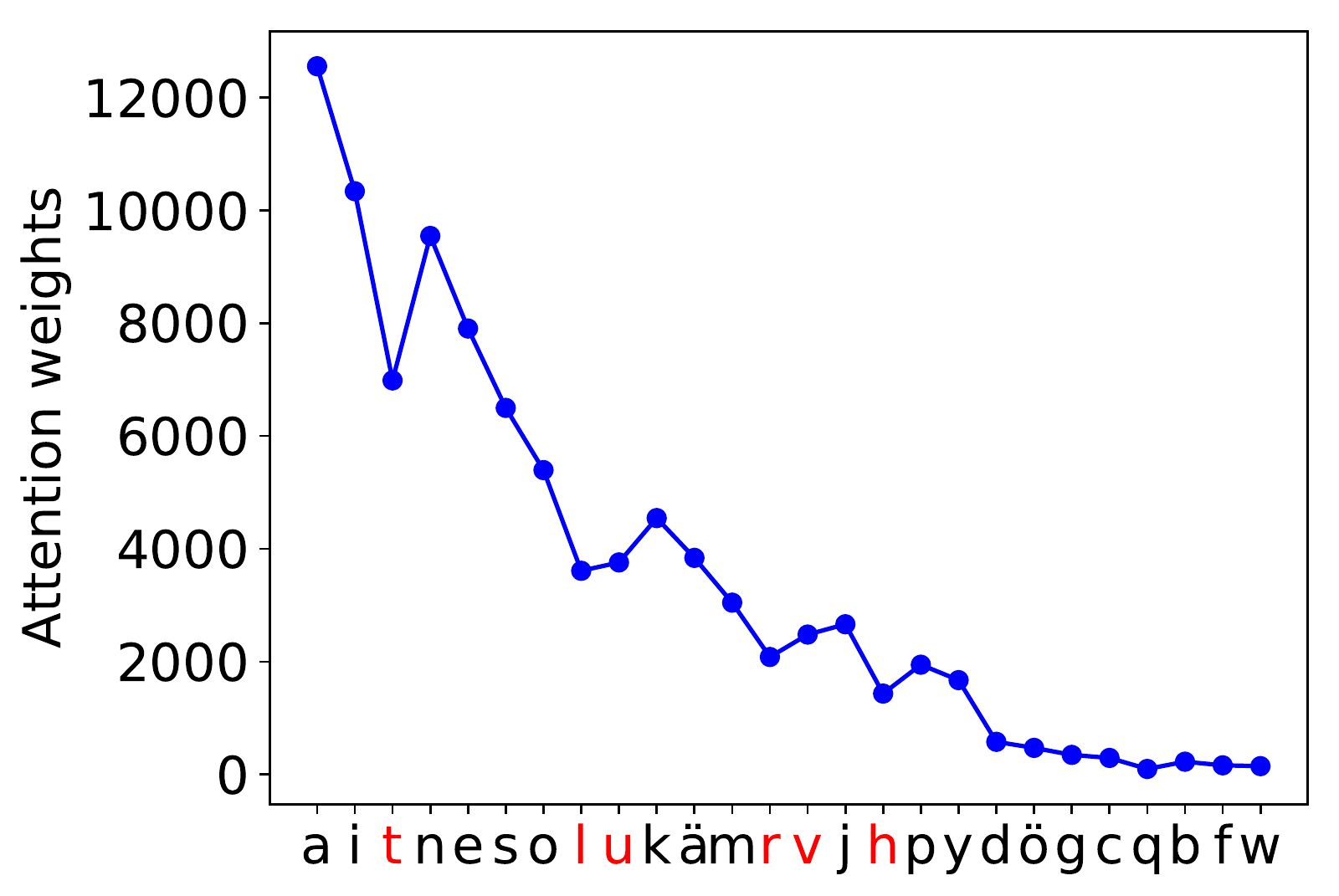}
    \caption{Attention distributions over characters in \textit{char-d7} trained on FI$\rightarrow$EN, summing up all the attention weights over each character. The characters are sorted in descending order of frequency and the characters in red are some exceptions in the trend curve.}
    \label{fig:att_sum_char}
    \end{center}
\end{minipage}%
    \hfill%
\begin{minipage}[t]{0.44\linewidth}
    \begin{center}
    \includegraphics[totalheight=4.4cm]{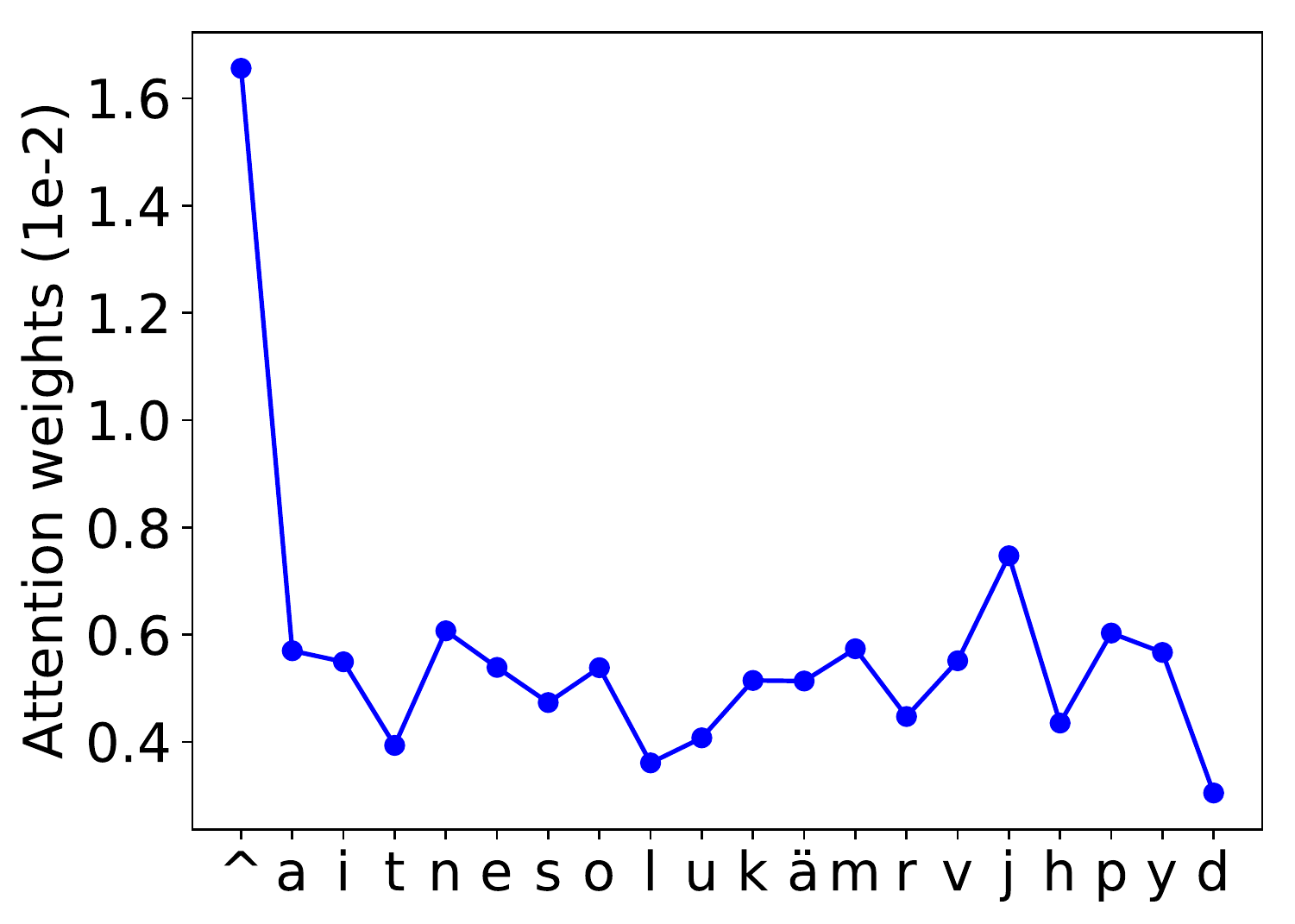}
    \caption{Attention weight over the 20 most frequent characters in \textit{char-d7} trained on FI$\rightarrow$EN, averaging all the attention weights over each character. ``\^{}'' denotes the separator.}
    \label{fig:att_avg_dis}
    \end{center}
\end{minipage} 
\end{figure}

\subsubsection{Effect of Encoder Depth}

\textit{char-d7} is superior to \textit{char-d4} in the translation task but it is still not clear how hidden states from NMT models with different encoder depths perform on the morphological probing tasks.  
As the hidden states in the last layer are fed to the decoder in the character-level NMT models, we first compare the hidden states from the last layer of both models, i.e. the 7th layer of \textit{char-d7} and the 4th layer of \textit{char-d4}. Figure~\ref{fig:d4d7last} displays the performance in all the morphological probing tasks. Generally, char-d7 outperforms char-d4 on the probing task, with the exception of the infinitive probing task. However, note that a comparison of the last encoder layer does not tell the full story. Looking at the evolution of probing performance over layers (Figure~\ref{fig:d4d7evolution}), we can see that \textit{char-d7} typically achieves the highest probing accuracy at the first layer, outperforming char-d4.

\section{Attention Mechanisms}
\label{sec:attention_distribution}

In the encoder-decoder attention \cite{bahdanau15joint,luong2015effective}, a higher attention weight means that the source token contributes more to the prediction at current step. Thus, we could utilize the attention distributions to explore how CHAR models pay attention to the source characters during translation.

CHAR models encode information into a longer sequence which essentially increases the representational capacity of the encoder. We are interested in exploring the effect of restricting the capacity of the encoder states that are passed to the decoder. Thus, we apply the word-level attention which attends to a character of each word and potentially enforces the character captures the full word information. 

\subsection{Attention Distributions over Characters}

We explore the attention distributions that are generated when translating \textit{newstest2015}. 
In addition to all the characters, we also consider the separator character. We calculate the attention weights over each source character. It is interesting that the sum of attracted attention is basically consistent with the frequency of characters in the source language (Finnish) which is shown in Figure~\ref{fig:att_sum_char}. This pattern indicates that most of the characters are treated equally during the overall decoding. However, when we average all the attention weights over each character and compare the separator with other characters, the separator apparently attracts much more attention as shown in Figure~\ref{fig:att_avg_dis}, which illustrates the attention weights over the 20 most frequent characters.  

\begin{table}[!t]
\centering
\begin{tabular}{ccc}
\toprule
Model&BLEU&Drop\\
\midrule
\addlinespace
char-d7&16.0&1.2 \\
bpe-d7&14.3&2.6 \\
\bottomrule
\end{tabular}
\caption{\label{table-word-attention} Results of applying word-level attention to both CHAR and BPE-based models. }
\end{table}

We also count the the frequency of attracting the most attention at each decoding step for all the source characters. We find that the separator accounts for 31.4\% of all the characters that have the highest weights. 
In addition, there are 29.6\% of normal characters and 39.6\% of separators in the target-side that distribute the highest attention weight to a source separator, which indicates that a large portion of normal characters also extract most of features from separators. 

The separators attract a lot of attention which is similar to the attention patterns of BERT \cite{devlin2019bert} found by \newcite{clark2019bert}. However, they find that the attention to the separators is used as a ``no-op'' for attention heads. In our settings, there is only one layer attention with one attention head, and we cannot regard the attention to separators as a ``no-op''.
Since the separators make the word representations better in morphology, we argue that the separators between words in the character sequences are encoded with rich linguistic features and contribute to the translation, which is different from the separators between sentences in BERT.

\subsection{Word-Level Attention}

As we have shown that separators have captured some linguistic knowledge, we enforce the word-level attention only attends to separators. 
We retrain the models with word-level attention from scratch. We also apply the word-level attention to BPE-based models as comparison. In that case, the attention attends to the last subword of a word. 

The BLEU scores of the models with word-level attention are given in Table~\ref{table-word-attention}. The BLEU scores drop in both \textit{char-d7} and \textit{bpe-d7}. It indicates that restricting the capacity of encoder states that are passed to the decoder has a negative effect on the BLEU scores.  The smaller drop on \textit{char-d7} means that the word-level information can be better extracted from the character-level hidden states compared to the subword-level hidden states. 
\textit{char-d7} does not perform too badly despite only attending to the separators. This indicates that the attention mechanism is very flexible and could force the model to encode more information into separators. 
However, since there is only one attention head, some information from the source is inevitably lost. It would be interesting to explore a multi-layer attention with multiple heads, such as Transformer \cite{vaswani2017Attention} attention, where the information could be extracted from the other attention heads as well. We leave this as future work. 

\section{Conclusion}

CHAR models have been shown to perform better than BPE-based models in NMT yet they pose new challenges for interpretability. In this paper, we investigate CHAR models via the WSD and six morphological probing tasks to learn how CHAR models learn word senses and morphology, in the case of translating Finnish into English. We also explore the attention distribution pattern and a sparse word-level attention to learn the working mechanism of attention. 

In the probing tasks, we find that separators also have captured some linguistic knowledge. 
We apply different composition methods to the characters of a word, and we demonstrate that the word sense and morphological information is distributed over all the characters rather than some specific characters. Moreover, characters at different positions play different roles in learning linguistic knowledge. 
CHAR models are better at learning morphology but we need a more complicated composition method, such as a randomly initialized LSTM, to extract all the encoded information. 
These results on probing tasks show that we can extract word sense information and morphological features from character-level hidden states and that these features are encoded in different ways. 
In addition, we explore the effect of encoder depth and show that CHAR models require more layers to encode word senses, which explains why only deeper CHAR models outperform BPE-based models. 
The attention distribution shows that separators attract a lot of attention, and we show that the sparse word-level attention only attending to separators is workable but not enough for translation. 

As we have shown that characters at different positions specialize in learning word senses and morphology, it will be interesting to explore sparse attention with multiple heads in the future which could learn to extract features from different aspects. 

\section*{Acknowledgments}
We thank all reviewers for their valuable and insightful comments. 
We acknowledge the computational resources provided by CSC in Helsinki and Sigma2 in Oslo through NeIC-NLPL (www.nlpl.eu). 
GT is mainly funded by the Chinese Scholarship Council (NO. 201607110016). 

\bibliography{references}
\bibliographystyle{acl}

\end{document}